\documentclass[10pt,twocolumn,letterpaper]{article}

\usepackage[pagenumbers]{cvpr} %

\usepackage{bibunits}

\usepackage[dvipsnames]{xcolor}

\definecolor{cvprblue}{rgb}{0.21,0.49,0.74}
\usepackage[pagebackref,breaklinks,colorlinks,citecolor=cvprblue]{hyperref}

\usepackage{graphicx}
\usepackage{smile_symbols}
\usepackage{bbding}

\def\logo{\makebox[0pt][l]{\hspace{-3pt}\raisebox{-0.3ex}{\includegraphics[height=16pt]{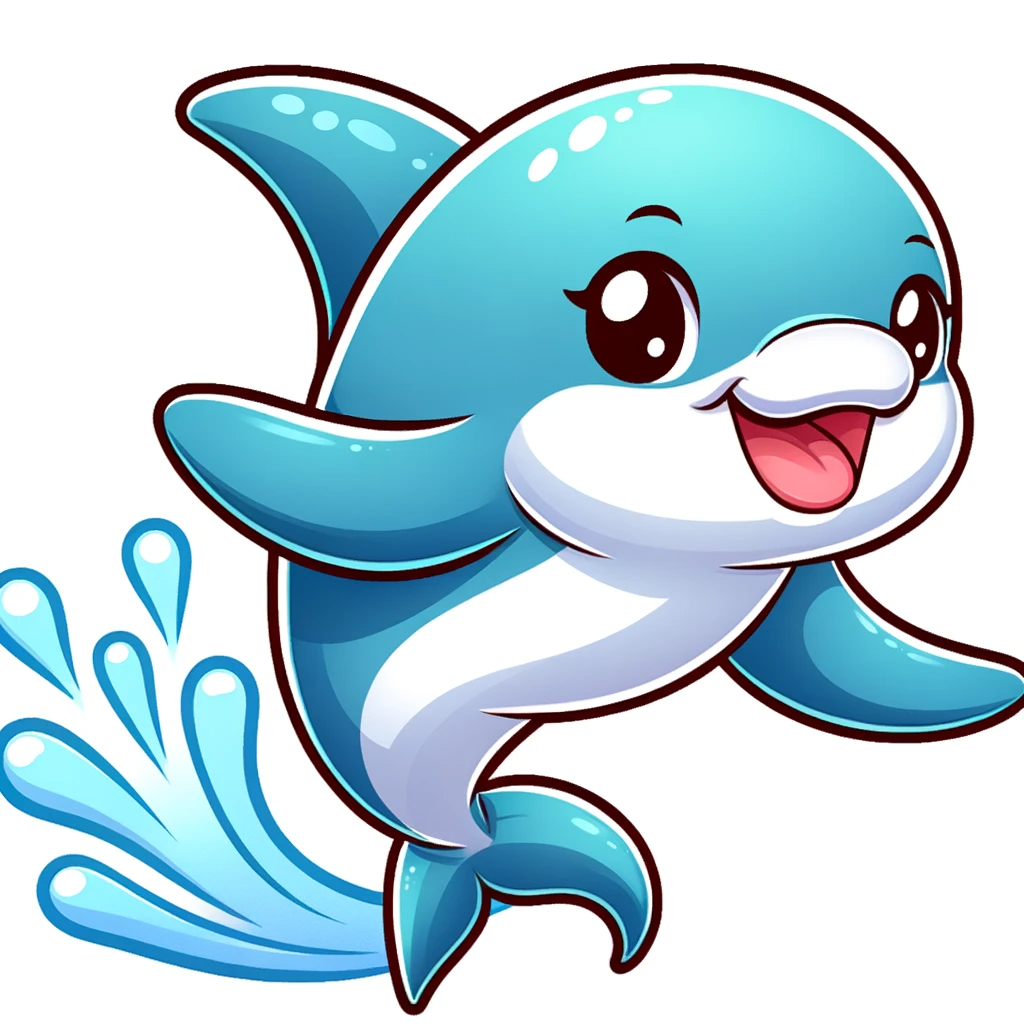}}}}

\newcommand{\ours}{\texttt{VaQuitA}\xspace}

\title{ \ours~\logo~~~~: Enhancing Alignment in LLM-Assisted Video Understanding}

\author{Yizhou Wang$^\dag\thanks{Work was done while Yizhou Wang was an intern at Adobe Research.}$, Ruiyi Zhang$^{\ddag}$, Haoliang Wang$^\ddag$, Uttaran Bhattacharya$^\ddag$, Yun Fu$^\dag$ and Gang Wu$^\ddag\thanks{Corresponding author.}$\\
$^\dag$Northeastern University, $^\ddag$Adobe Research \\
wyzjack990122@gmail.com, ruizhang@adobe.com, hawang@adobe.com, utta005@gmail.com, \\ yunfu@ece.neu.edu, gawu@adobe.com
}

\begin{document}
\maketitle

\begin{abstract}
Recent advancements in language-model-based video understanding have been progressing at a remarkable pace, spurred by the introduction of Large Language Models (LLMs). However, the focus of prior research has been predominantly on devising a projection layer that maps video features to tokens, an approach that is both rudimentary and inefficient. In our study, we introduce a cutting-edge framework, \ours, designed to refine the synergy between video and textual information. At the data level, instead of sampling frames uniformly, we implement a sampling method guided by CLIP-score rankings, which enables a more aligned selection of frames with the given question. At the feature level, we integrate a trainable Video Perceiver alongside a Visual-Query Transformer (abbreviated as VQ-Former), which bolsters the interplay between the input question and the video features. We also discover that incorporating a simple prompt, ``Please be critical'', into the LLM input can substantially enhance its video comprehension capabilities. Our experimental results indicate that \ours consistently sets a new benchmark for zero-shot video question-answering tasks and is adept at producing high-quality, multi-turn video dialogues with users.%
\end{abstract}
\vspace{-1.5em}    
\section{Introduction}~\label{sec:intro}
\begin{figure}[tb]
  \centering
  \includegraphics[width=0.9\linewidth]{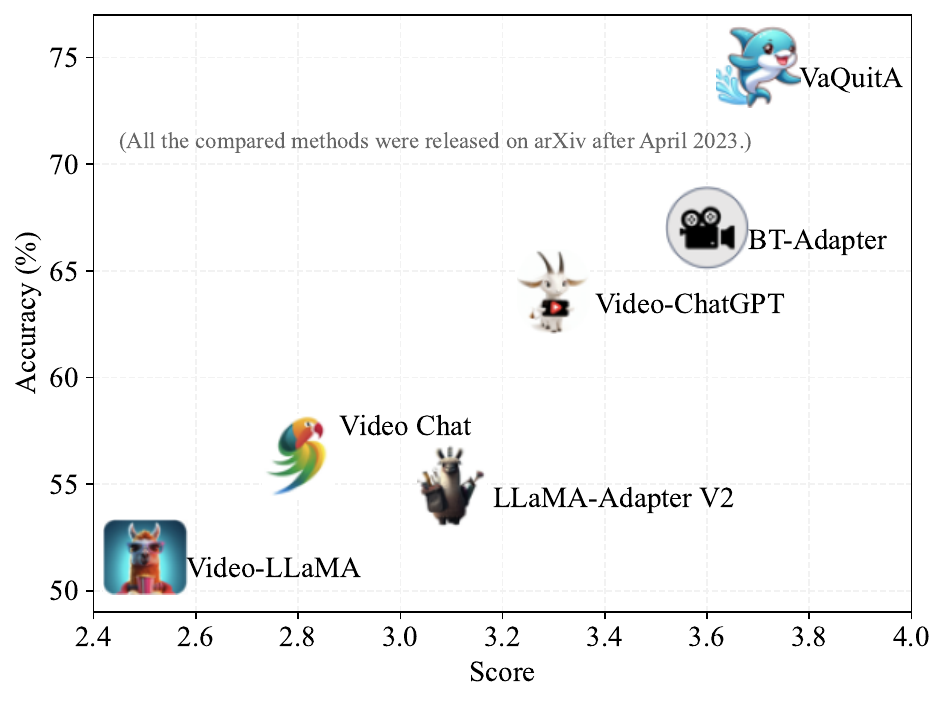}
  \vspace{-1em}
  \caption{Accuracy ($\uparrow$) and relative score ($\uparrow$) comparison on MSVD-QA dataset of current state-of-the-art LLM-based zero-shot video understanding models (evaluated using GPT-3.5-turbo API). \ours achieves the best performance in both evaluation metrics. Please check more details in Sec.~\ref{sec: exp-zsvqa}.}
  \label{fig: first}
  \vspace{-2em}
\end{figure}
The rise of deep learning tools for video interpretation has ushered in significant progress in video-centric tasks~\cite{xu2021videoclip,wqb+22,wglf23}. Yet, current models for video comprehension often falter when engaging in spontaneous discussions about video content~\cite{zhong2022video}. A dialogue system rooted in video content can transform video searches, enhance monitoring techniques, and assist in summarizing pivotal events. Importantly, it offers a unified, accessible interface for video tasks, including action recognition, location identification, detection, retrieval, and tracking~\cite{mu2023embodiedgpt}. This proficiency is especially noteworthy, highlighting the model's ability to understand temporal and spatial indications, grasp context, and perceive extended relationships~\cite{liu2023interngpt}.

Existing research in Large Video Language Models~\cite{yang2022zero,zhang2023video,gao2023llama,li2023videochat,maaz2023video,liu2023one} predominantly adopts a uniform sampling strategy for frame selection. These models typically use a single projection layer to transfer and align video semantic content into the token space. The resulting tokenized video embeddings are then concatenated with query embeddings and fed into Large Language Models (LLMs) for response generation. However, this straightforward approach overlooks the need to guide the projection of video features into specific text representations and to discern which aspects of the video, whether spatial or temporal, should be emphasized. Given the constraints of limited training data, this methodology often leads to suboptimal performance in out-of-distribution video understanding tests~\cite{maaz2023video}. Moreover, in practical, real-world scenarios, this can lead to perplexing errors in video conversation systems~\cite{liu2023one}.

To mitigate the above problems, we introduce \ours, an innovative framework that redefines the approach to video and textual information integration. \ours diverges from traditional methodologies by implementing a CLIP~\cite{radford2021learning}-score guided frame sampling method. This innovation allows for the selection of frames that exhibit a higher relevance to the input question, thereby addressing the limitations of uniform frame sampling. The framework further advances the interaction between video content and textual queries through the integration of a trainable Video Perceiver. This component enhances the processing of video features, ensuring a more nuanced understanding of the visual content. Complementing this is our Visual-Query Transformer (VQ-Former), which acts as a pivotal element in aligning the video features with the textual query, facilitating a more coherent and context-aware interplay. Furthermore, \ours incorporates a novel approach in its interaction with LLMs. By introducing a simple, yet effective prompt --- ``Please be critical'' --- into the LLM input during testing, we notice a marked enhancement in the model's capability to understand and interpret video material. This refinement leads to a more critical and discerning analysis by the LLM, enhancing its performance in complex video understanding tasks. %

In summary, the main contributions of the paper are:
\begin{itemize}
    \item \ours, a novel video conversation model that strengthens the alignment of text features and video features. The alignment lies in both the raw data level and the feature level, which enhances the fusion of question and video information, leading to stronger reasoning ability of the Video Question Answering model.
    \item We uncover the fact that adding an additional prompt, ``Please be critical'', before the question can improve the understanding ability of \ours universally.
    \item Our proposed \ours achieved state-of-the-art performance on the Zero-shot Video Question Answering task. It can also conduct top-notch multi-turn conversations.
\end{itemize}

\section{Related Works} 
We briefly summarize existing works in the related areas of video conversation, vision large lanugage models, and visual-text alignment.

\paragraph{Video Conversation.}
With the rapid development of LLMs, researchers begin to transfer their extraordinary reasoning abilities to the video conversation area. VideoChat~\cite{li2023videochat} integrates foundational video models and LLM using a learnable neural interface, comprising two branches: VideoChat-Text which textualizes videos in real-time, and VideoChat-Embed which encodes video into embeddings using Video Foundation Models and Token Projection; the processed video content and questions are then passed to the LLM for generating answers. Video-LLaMA~\cite{zhang2023video} employs a multi-branch cross-modal pre-training approach, effectively achieving alignment between vision-language and audio-language. Nevertheless, both these two approaches have limited ability to handle long videos and have no quantitative results. Different from these, VideoLLM~\cite{chen2023videollm} focuses on video sequence understanding, handling eight distinct video understanding tasks across four datasets, including online action detection, highlight detection, moment query, and natural language query, among others. The system employs a modality encoder, semantic translator, decoder-focused reasoning unit, and straightforward task heads, centralizing LLM for seamless reasoning between video and text sequences; the semantic translator aligns visual and text encoding, making VideoLLM state-of-the-art in many tasks by a substantial degree. More recently, Branching Temporal Adapter (BT-Adapter)~\cite{liu2023one} method extends image-language pretrained models into the video domain by acting as a plug-and-use temporal modeling branch next to the pretrained visual encoder, which is fine-tuned with the main backbone remaining unchanged. %
Despite the progress, the current video conversation capability is still limited, and none of the existing models can answer half of the questions correctly in the zero-shot setting on Activity Net-QA~\cite{yu2019activitynet} dataset so far. 

\begin{figure*}[tb]
  \centering
  \includegraphics[width=.9\linewidth]{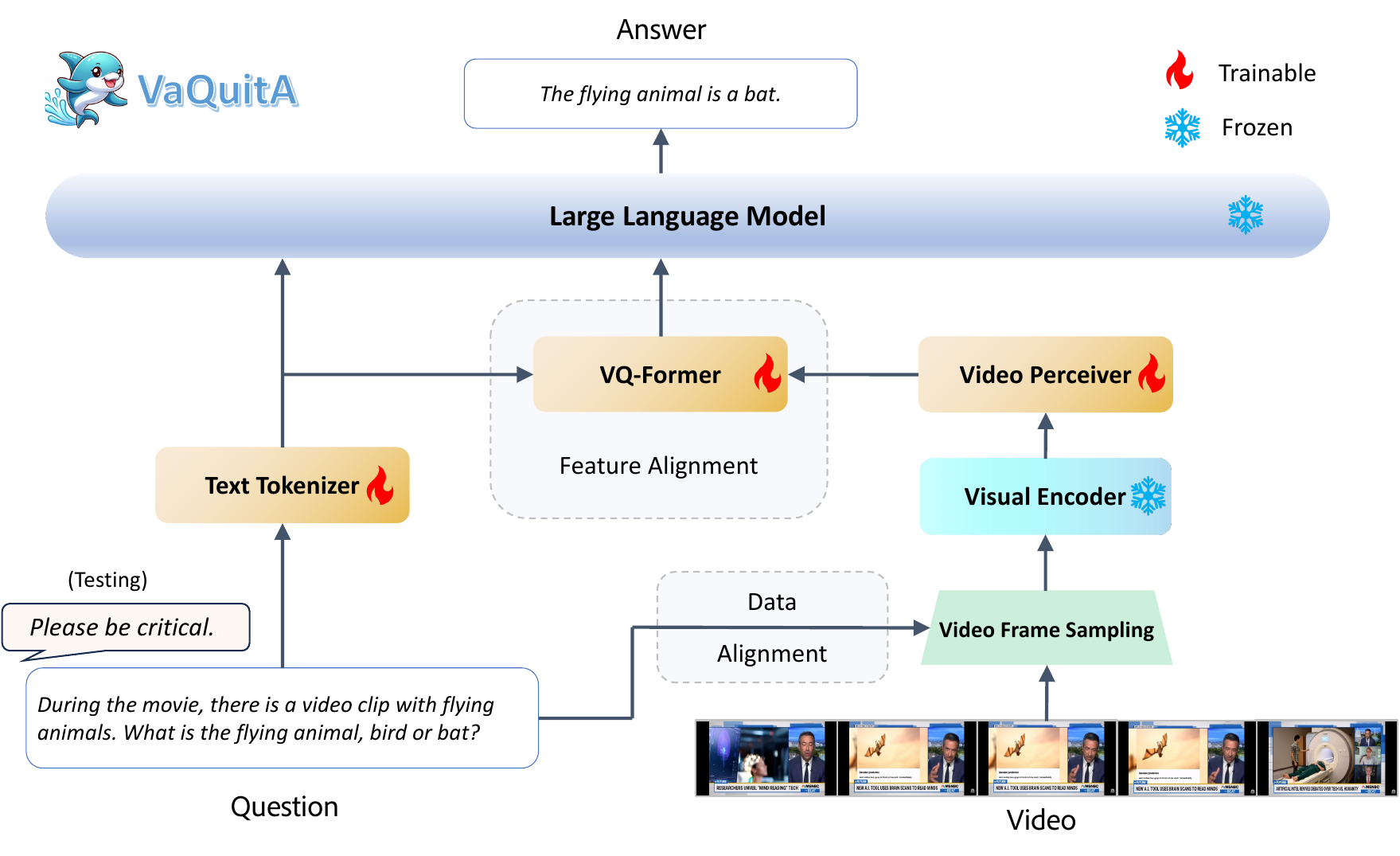}
  \vspace{-1em}
  \caption{{\bf Framework overview.} In response to a specific question, our framework begins by processing the input video with a sampling module that identifies key frames based on their relevance to the question's context. These frames are then processed by a pre-trained visual encoder to obtain spatio-temporal features. These features are subsequently refined into condensed embeddings by our newly developed Video Perceiver. In parallel, the question undergoes tokenization. Both the video and text embeddings are then synergized using our Visual-Query Transformer, which aligns the multimodal information more effectively. The resulting text-influenced video features are concatenated with the text embeddings and fed into the Large Language Model to generate the answer. During the testing phase, we propose to add an additional prompt, ``Please be critical'', before the question for performance enhancement. The whole proposed \ours framework supports end-to-end training.  Best viewed in color.}
  \label{fig: framework}
  \vspace{-1em}
\end{figure*}

\paragraph{Vision Large Language Models.}
Recent progress in computer vision has been propelled by the emergence of groundbreaking vision-language models. These models mark a considerable step forward in developing versatile vision models that can handle multiple tasks at once~\cite{gupta2022towards,maaz2022class}. A standout model in this realm is CLIP~\cite{radford2021learning}, trained on 400 million image-text pairs, showcasing exceptional zero-shot capabilities across many benchmarks. In more recent times, Flamingo~\cite{alayrac2022flamingo} is a new family of Visual Language Models designed to rapidly adjust to novel tasks using a minimal number of annotated examples. It proposes perceiver resampler and gated cross-attention architectures, enabling its superior few-shot learning capabilities by training on large-scale multimodal web datasets with mixed text and images. BLIP-2~\cite{li2023blip} represents an effective method for pre-training that leverages existing image encoders and language models that have undergone pre-training, connecting them using a lightweight Querying Transformer in two stages: the vision-language representation and vision-to-language generative training. While some of these models are compatible with both images and videos, there is a growing demand for a robust, video-specific large vision-language model.

\paragraph{Visual-Text Alignment.}
Contemporary progress in aligning visual-text features primarily revolves around the concept of harmonizing multimodal features originating from various representational spaces. The foundational work~\cite{Duan_2022_CVPR} highlighted the challenges of aligning evolving features during training. Progressing from this, the Multi-Modality Cross Attention Network~\cite{Wei_2020_CVPR} and the ``MVPTR'' framework~\cite{li2022mvptr} both emphasized the significance of fine-grained feature alignment and cross-modal interactions, illustrating a shift towards more sophisticated semantic alignment tasks. Further innovations in multimodal fusion were proposed through CentralNet~\cite{vielzeuf2018centralnet} presenting a multilayered integration approach, and ADAPT~\cite{lin2022adapt} which introduced dynamic action-based context alignment for Vision-Language Navigation, showcasing the practical application of alignment in autonomous systems. These developments culminated in the Multimodality-guided Visual Pre-training (MVP) approach~\cite{wei2203mvp}, which leveraged large-scale image-text datasets to refine the alignment process, marking a significant step forward in pre-training methodologies. Contrasting with existing approaches, our method enforces the alignment of video and text embeddings through a novel video feature resampling network and a bespoke cross-attention module tailored for the LLM input space. This approach represents an innovative direction in the field of large vision language modeling.

\section{\ours Framework}

Our proposed \ours framework consists of three novel components: Data Alignment module (Sec.~\ref{sec: da}), Feature Alignment module (Sec.~\ref{sec: fa}), and test-time Prompt Engineering (Sec.~\ref{sec: pe}). The entire pipeline is illustrated in Fig.~\ref{fig: framework}.

\subsection{Data Alignment}\label{sec: da}
Existing methodologies typically employ a uniform sampling approach to extract frames for video conversation~\cite{maaz2023video,bhattacharya2023video,yang2023vid2seq} or video understanding in general~\cite{li2022uniformerv2, wang2022internvideo}. Such uniform sampling method, while straightforward, often results in the loss of critical information contained in the frames that are not selected, affecting the model's ability to understand videos effectively. To address this limitation, we present a new method in our \ours that leverages the semantic similarity between the video frames and the question prompt for frame selection. This technique ensures a more congruent alignment between the features of the question and those of the frames at the raw data level. We refer to this as the ``Data Alignment'' module.

\begin{figure}[tb]
  \centering
  \includegraphics[width=0.9\linewidth]{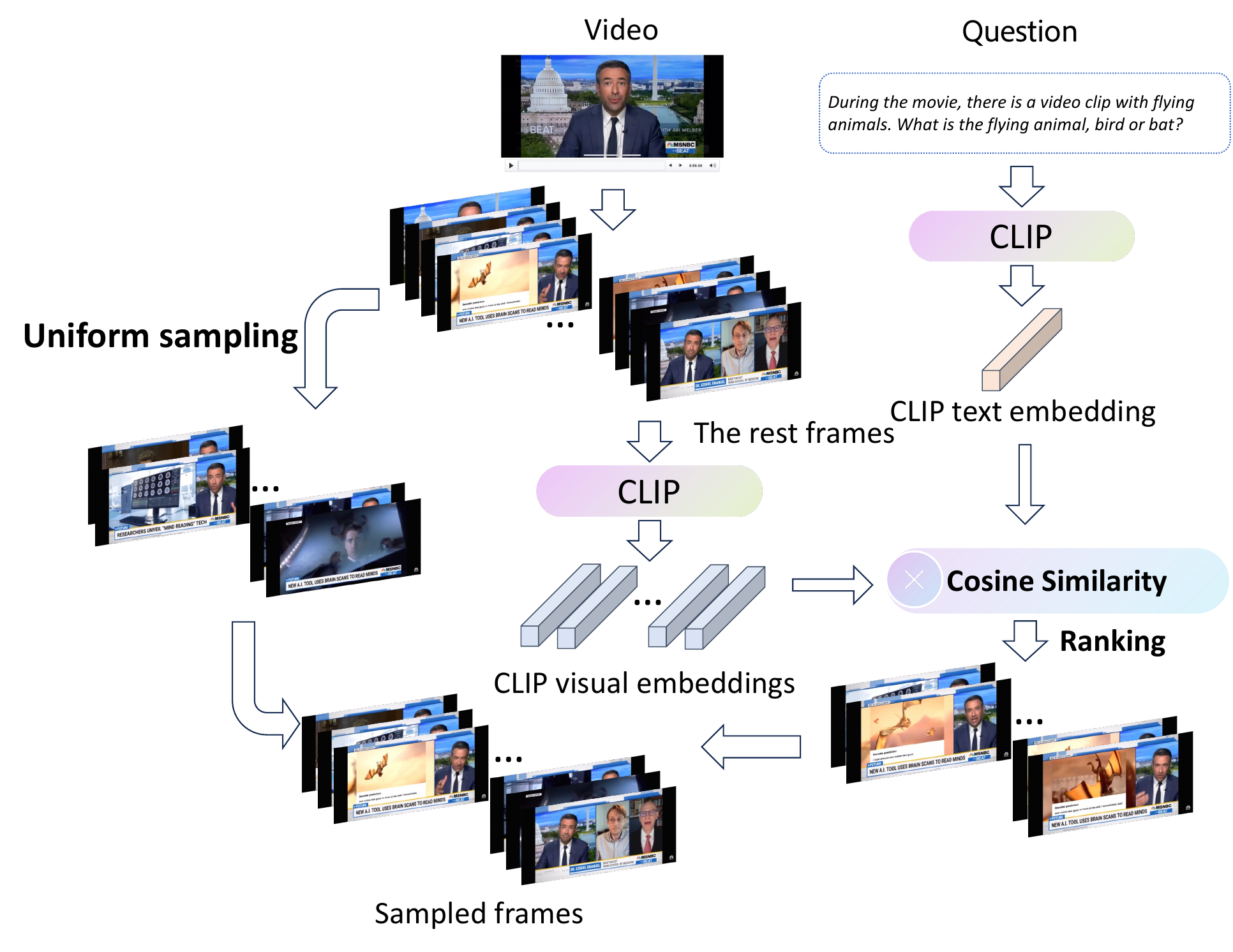}
  \vspace{-1em}
  \caption{{\bf Data alignment.} Our proposed sampling module consists of both uniform sampling and similarity-based sampling for the training process. Best viewed in color.}
  \label{fig: data_alignment}
  \vspace{-1em}
\end{figure}

\paragraph{CLIP Feature Similarity-based Frame Selection for Training.}
Given the input video of $L$ frames in total, instead of getting a certain number of frames with only uniform sampling, we also select frames based on the similarity between the frame features and the input query. Suppose we sample $T$ frames in total, we propose to select $\frac{T}{2}$ frames uniformly over the temporal dimension and another $\frac{T}{2}$ frames using the similarity-based approach. Specifically, we extract the text feature of the query using CLIP model, denoted as $f_{\text{query}}$, and the visual features of the remaining frames that are not selected as $\{f_{\text{video}}^1, f_{\text{video}}^2, \cdots, f_{\text{video}}^{L-\frac{T}{2}}\}$, the similarity is calculated as 
\begin{equation}
    \text{Cosine-Similarity}(f_{\text{query}}, f_{\text{video}}^i) = \frac{f_{\text{query}} \cdot f_{\text{video}}^i}{\norm{f_{\text{query}}}_2 \times \norm{f_{\text{video}}^i}}_2,
\end{equation}
and we select the indices of the top $\frac{T}{2}$ values. The motivation is that uniform sampling will lead to information loss due to its non-adaptivity, and by employing the proposed similarity-based approach, frames that are most related to the question will be selected, improving representation learning ability. The schematic diagram is in Fig.~\ref{fig: data_alignment}. 

\paragraph{Uniform Sampling for Testing.}Our proposed sampling method is implemented during the training phase of our model. For the testing or inference stage, we revert to uniform sampling due to efficiency considerations and the need for speed in real-world applications. This approach is enough to demonstrate satisfactory performance in our experimental evaluations. We supplement an example in the supplementary showing that our sampling approach can improve testing performance as well compared with uniform sampling, despite being slower.

\subsection{Feature Alignment}\label{sec: fa}

\begin{figure}[tb]
  \centering
  \includegraphics[width=0.9\linewidth]{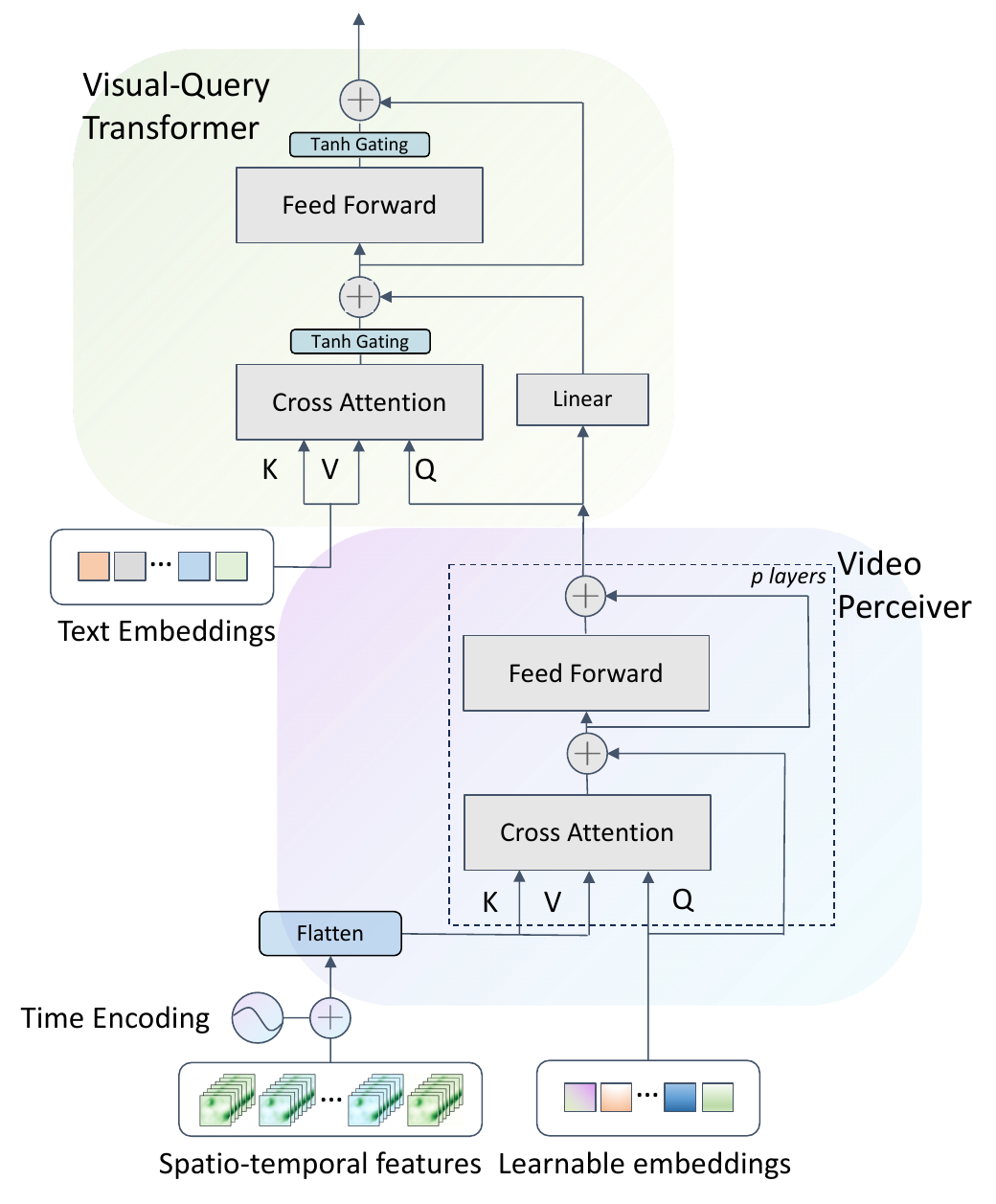}
  \vspace{-1em}
  \caption{{\bf Feature alignment}. The extracted spatio-temporal features of the video clip first go through Video Perceiver for representative embedding extraction, and are afterwards sent to Visual-Query Transformer for interleaving with text embeddings.}
  \label{fig: feature_alignment}
  \vspace{-1em}
\end{figure}

Visual data are regarded as the reflection and capture of the physical world while text data can be seen as the abstract of the understanding of the world and the fundamental logic~\cite{lecun2022auto}. The successful alignment of visual and textual information is significant for an intelligent system to work appropriately. Instead of directly concatenating the tokenized text and visual features to put into LLM~\cite{liu2023visual,maaz2023video,zhang2023video,chen2023videollm,liu2023one}, we propose a novel Visual-Query Transformer, abbreviated as VQ-Former, to produce text-guided video embeddings before concatenation with the text embeddings. The inspiration comes from recent work on visual-text pretraining~\cite{li2023blip,alayrac2022flamingo}, and the illustration of the architecture is provided in Fig.~\ref{fig: feature_alignment}.

\subsubsection{Video Perceiver}
Given a sampled video snapshot, we first apply a pretrained CLIP model to extract semantic features for each frame. Suppose the extracted spatio-temporal feature embeddings are $F \in \RR^{T\times n\times d}$, where $T$ is the sampled frame number, $n$ is the number of features for each frame, \ie, the patch number for CLIP model, and $d$ denotes the dimension of feature. To facilitate the alignment with text embeddings and input into the LLM, we need to resample and reduce the number of video features for computation feasibility. Inspired by Perceiver Resampler~\cite{alayrac2022flamingo}, we put forward Video Perceiver which transforms the spatio-temporal visual attributes into a number of learned output tokens. The spatio-temporal features are first added by Time Encodings of shape $T\times 1\times d$ to store the sequence order information and then flattened to shape $Tn \times d$ for the cross-attention module dimension match. This cross-attention module employs a collection of learned latent vectors to query (Q), while the keys (K) and values (V) combine the flattened spatio-temporal visual attributes with these learned latent vectors. The shape of the learned latent embeddings is $m \times d$, where $m$ denotes the number of latent embeddings. The weights of the learned latent embeddings are randomly initialized. Following transformer~\cite{vaswani2017attention}, feed-forward networks and residual connections are added for efficient modeling in our Video Perceiver. The output embedding shape remains the same as the input learnable embeddings, \ie, $m\times d$. We denote the number of layers of Video Perceiver as $p$.

\subsubsection{Visual-Query Transformer}
The input question goes through a learnable text tokenizer and turns into query embeddings, which, together with the learnable embeddings output of the video perceiver, are sent into VQ-Former. The layers derive their keys and values from vision features, whereas the queries originate from the language inputs. Visual-Query Cross-Attention layer is applied for the query feature (denoted as $X \in \RR^{l \times d_{\text text}}$) and video feature (denoted as $M\in \RR^{m\times d}$) interleaving, where $l$ is the length and $d_{\text text}$ is the text embedding dimension.
\paragraph{Visual-Query Cross Attention.} In the Visual-Query Cross Attention layer, we adopt a multi-head mechanism as in Transformer~\cite{vaswani2017attention}. We denote the head index as $h$ and the inner feature dimension of each head as $d_h$. Given the input learned video feature, we first apply Layer Normalization~\cite{ba2016layer} and denote the normalized one as $M$. Then we have the $Q,K,V$s for each head calculated as
\begin{align}
    Q^{(h)} &= \frac{M W_Q^{(h)}}{s_q} , \\
    K^{(h)} &= X W_{K}^{(h)}, \\
    V^{(h)} &= X W_{V}^{(h)},
\end{align}
and
\begin{equation}
    O_a^{(h)} = \texttt{Softmax}(Q^{(h)}{K^{(h)}}^\top ) V^{(h)} W_O^{(h)},
\end{equation}
where $W_Q^{(h)} \in \RR^{d \times d_h }$, $W_{K}^{(h)} \in \RR^{d_{\text text} \times d_h }$,  $W_{V}^{(h)} \in \RR^{d_{\text text} \times d_h }$ and $W_O^{(h)} \in \RR^{ d_h \times d_{\text text}}$ are leanable weight parameters of head $h$. $s_q$ is a scaler representing the scale parameter, and $h$ represents the head index. Denoting the Visual-Query Cross Attention layer output as $O_a$, we have that $O_a$ is the multi-head concatenation of each head output:
\begin{equation}
O_a = \text{Concat}(O_a^{(1)},\cdots, O_a^{(H)}),
\end{equation}
where $H$ denotes the head number. The dot product attention computation aligns the semantics of the video embedding $M$ and query embedding $X$, contributing to the selection and learning of the visual features more relevant with the question. The multi-head design enables the exploration of the weight parameters in more feature subspaces for superior representation learning~\cite{vaswani2017attention}.%

\paragraph{VQ-Former Overview.} We use \texttt{Cross\_Attn} to denote the Visual-Query Cross Attention, and the entire procedure of ourVisual-Query Transformer can be written as: 
\begin{align}
O_a &= \texttt{Cross\_Attn} (M, X), \\
M' &= O_a \cdot \tanh(g_{\text{attn}}) + MW_M, \\
O_f &= \texttt{Feed\_Forward}(M'), \\
M'' &= O_f \cdot \tanh(g_{\text{ff}}) + M'.
\end{align}
Here the learnable parameter $W_M \in \RR^{d \times d_{\text text}}$ is applied to transform the dimension of the video representative features into token feature dimension for the residual architecture. $\tanh$ denotes Hyperbolic Tangent function and \texttt{Feed\_Forward} denotes a Feed Forward net block containing $2$ linear layers with Layer Normalization and GELU~\cite{hendrycks2016gaussian} activation layer. $g_{\text{attn}}$ is Attention Gate and $g_{\text{ff}}$ is FeedForward Gate, which are both learnable scalar parameters borrowed from Flamingo~\cite{alayrac2022flamingo} for improved stability and performance.  Eventually, the output question-interacted video features $M''$ are input to the LLM together with the input question embeddings. Different from the existing visual-text interleaving architectures, \eg, Q-Former~\cite{li2023blip} or Gated Cross-Attention layer~\cite{alayrac2022flamingo}, our QV-Former converts visual features to Queries and text features to Keys and Values for attention value computation. The underlying rationale of our approach is to utilize the information from the query as a directive to enhance the learning of pivotal visual embeddings. This represents a shift from the conventional method, where visual information typically serves as a guide for the learning of text representations.

\subsection{End-to-end Training} \label{sec: e2e-training}
Our \ours supports end-to-end training: the trainable parameters include the Text Tokenier, the VQ-Former, and the Video Perceiver. The visual encoder (CLIP) and the Large Language model weights are derived from pretrained weights and are frozen during our training. The CLIP model employed to extract $f_{\text query}$ and $f_{\text video}$ is also frozen during the training. We employ the standard smoothed Negative Log-Likelihood Loss in NLP literature.

\begin{table*}[ht]
\centering
\setlength{\tabcolsep}{6pt}
\renewcommand{\arraystretch}{1.2}
\begin{tabular}{l c c c c c c}
\hline
\textbf{Model} & \multicolumn{2}{c}{\textbf{MSVD-QA}} & \multicolumn{2}{c}{\textbf{MSRVTT-QA}} & \multicolumn{2}{c}{\textbf{Activity Net-QA}} \\
\cline{2-7}
 & \textbf{Accuracy ($\uparrow$)} & \textbf{Score ($\uparrow$)} & \textbf{Accuracy($\uparrow$)} & \textbf{Score($\uparrow$)}  & \textbf{Accuracy($\uparrow$)} & \textbf{Score ($\uparrow$)} \\
\hline
\hline
FrozenBiLM$^*$~\cite{yang2022zero} & 32.2 & -- & 16.8 & -- & 24.7 & -- \\
VideoLLaMA$^\dag$~\cite{zhang2023video}  & 51.6 & 2.5 & 29.6 & 1.8  & 12.4 & 1.1 \\
LLaMA-Adapter$^\dag$~\cite{gao2023llama} & 54.9 & 3.1 & 43.8 & 2.7 & 34.2 & 2.7\\
Video Chat$^*$~\cite{li2023videochat} & 56.3 & 2.8 & 45.0 & 2.5 & 26.5 & 2.2 \\
Video-ChatGPT$^*$~\cite{maaz2023video} & 64.9 & 3.3 & 49.3 & 2.8 & 35.2 & 2.7 \\
BT-Adapter$^\dag$~\cite{liu2023one} & 67.0 & 3.6 & 51.2 & 2.9 & 46.1 & 3.2 \\
{\ours~(Ours)} & \textbf{74.6} & \textbf{3.7} & \textbf{68.6} & \textbf{3.3}  & \textbf{48.8} & \textbf{3.3} \\
\hline
\end{tabular}
\caption{Zero-Shot question-answering performance comparison of \ours with other models. Our \ours demonstrates SOTA performance across all examined datasets.$^*$ denotes the results reported in~\cite{maaz2023video} and $^\dag$ denotes the results reported in~\cite{liu2023one}.}
\label{tab: main-results}
\end{table*}

\subsection{Prompt Engineering} \label{sec: pe}
Prompt engineering~\cite{wei2022chain,zhou2022large,gu2023systematic} refers to the systematic design and modification of input prompts to guide machine learning models, particularly pretrained LLMs, to produce desired or more accurate outputs. The essence of this technique is rooted in the understanding that the input provided to a model doesn't merely serve as a query but also as a form of soft guidance, potentially shaping the model's behavior and outputs. In our experiments, we are excited to discover that in the testing phase, if we add a prompt ``Please be critical'' before the question, zero-shot question answering performance can be significantly and consistently improved. This might imply an intriguing principle that, unlike in question answering in NLP the models are demanded to be calmer or more organized~\cite{kojima2022large,yang2023large},  the model needs to be more critical or judgmental for video question answering tasks. An ablation study on the prompts is carried out in Sec.~\ref{sec: ablation-hyper} which verifies the implication.

\section{Experiments}
In the experimental implementation, we employ Llama 2 (7B)~\cite{lamma2_23} as the foundational LLM backbone and initialize its weight using the weights of LLaVA-1.5~\cite{liu2023improvedllava}. We fine-tune the trainable parameters in \ours using the video instruction dataset VideoInstruct-100K\footnote{\url{https://huggingface.co/datasets/MBZUAI/VideoInstruct-100K}}~\cite{maaz2023video}, comprising roughly 100,000 pairs of video instructions. The fine-tuning phase spans three epochs, utilizing a step size of value $2e-5$ and a total batch size of value $32$. For fair comparison, we keep the data-level hyperparameters as the same in literature: $T=100, d=1024, d_{\text text}=4096, $. We employ the ``clip-vit-large-patch14'' CLIP version for video feature extraction. Specifically, for the sampling-period features $f_{\text query}$ and $f_{\text video}$, we use the last layer of the CLIP model with dimension 768. For the video feature extraction before the video perceiver, we utilize the last but one layer of CLIP with patch number $n=256$ and feature dimension $d=1024$. We choose the number of learnable visual embeddings in Video Perceiver as $m=356$, which is the same as the dimension after spatio-temporal pooling in Video-ChatGPT~\cite{maaz2023video} for a fair comparison. The perceiver depth is set as $p=1$. For all the attention blocks in both Video Perceiver and VQ-Former, we set $d_h=64, H=8$ and scale parameter $s_q=8$. All the training experiments are conducted on eight A100 80GB GPUs. For testing, one GPU with $15$ GB GPU memory is sufficient for our model.
\subsection{Zero-shot Video Question Answering}\label{sec: exp-zsvqa}

We carry out an exhaustive quantitative assessment using several prevalent open-ended video question-answer datasets, encompassing MSRVTT-QA~\cite{xu2017video}, MSVD-QA~\cite{xu2017video}, and Activity Net-QA~\cite{yu2019activitynet}. Following~\cite{maaz2023video}, the assessments are performed in a zero-shot setting, utilizing GPT-guided evaluation to gauge the model's proficiency. This assessment method calculates the precision of the model’s predicted outputs (accuracy) and ranks them on a 1-5 scale (score). To ensure a fair comparison with the baselines, we employ Azure GPT-3.5-turbo API (March version) for evaluation, which is consistent with~\cite{maaz2023video}. Our \ours's efficacy is juxtaposed with other notable models, namely FrozenBiLM~\cite{yang2022zero}, VideoLLaMA~\cite{zhang2023video}, LLaMA-Adapter~\cite{gao2023llama}, Video Chat~\cite{li2023videochat}, Video-ChatGPT~\cite{maaz2023video} and BT-Adapter~\cite{liu2023one}. From Table~\ref{tab: main-results}, we can draw the conclusion that \ours performs consistently the best on Activity Net-QA dataset, outperforming the second best by a considerable margin on both accuracy and score evaluation. Compared to the current best models, our proposed \ours achieves $7.6\%$, $17.4\%$, $2.7\%$ gain in accuracy, and $0.1$, $0.4$, $0.1$ gain in score on MSVD-QA, MSRVTT-QA and Activity Net-QA datasets respectively.

\subsection{Multi-Round Conversation}\label{sec: multi-round}
\begin{figure*}[tb]
  \centering
  
  \begin{subfigure}{\linewidth}
    \centering
    \includegraphics[width=0.85\linewidth]{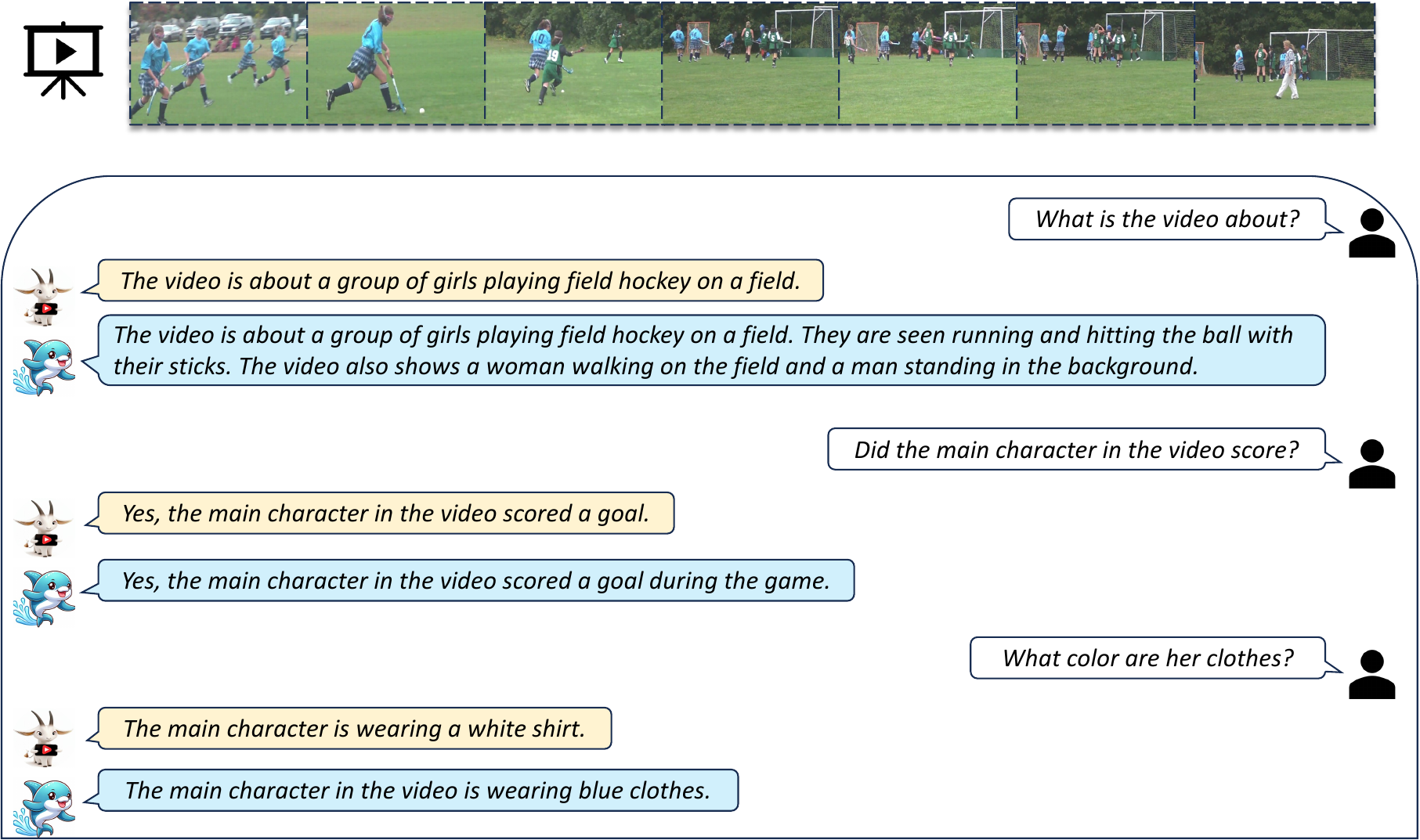}
    \caption{Given a video clip on a group of girls playing field hockey, we ask questions on the content of the video, whether the main character scores, and the color of the main character's clothes. Our \ours can answer all the questions correctly while the baseline Video-ChatGPT~\cite{maaz2023video} fails to tell the correct color of the clothes the girl is wearing. In addition, the generated answers of \ours are more detailed and specific like a human chatting with the user, while the responses of Video-ChatGPT are short and like being forced to complete a task. Best viewed in color.}
    \label{fig: demo_hockey}
  \end{subfigure}
  \vspace{0.2cm} %

  \begin{subfigure}{\linewidth}
    \centering
    \includegraphics[width=0.85\linewidth]{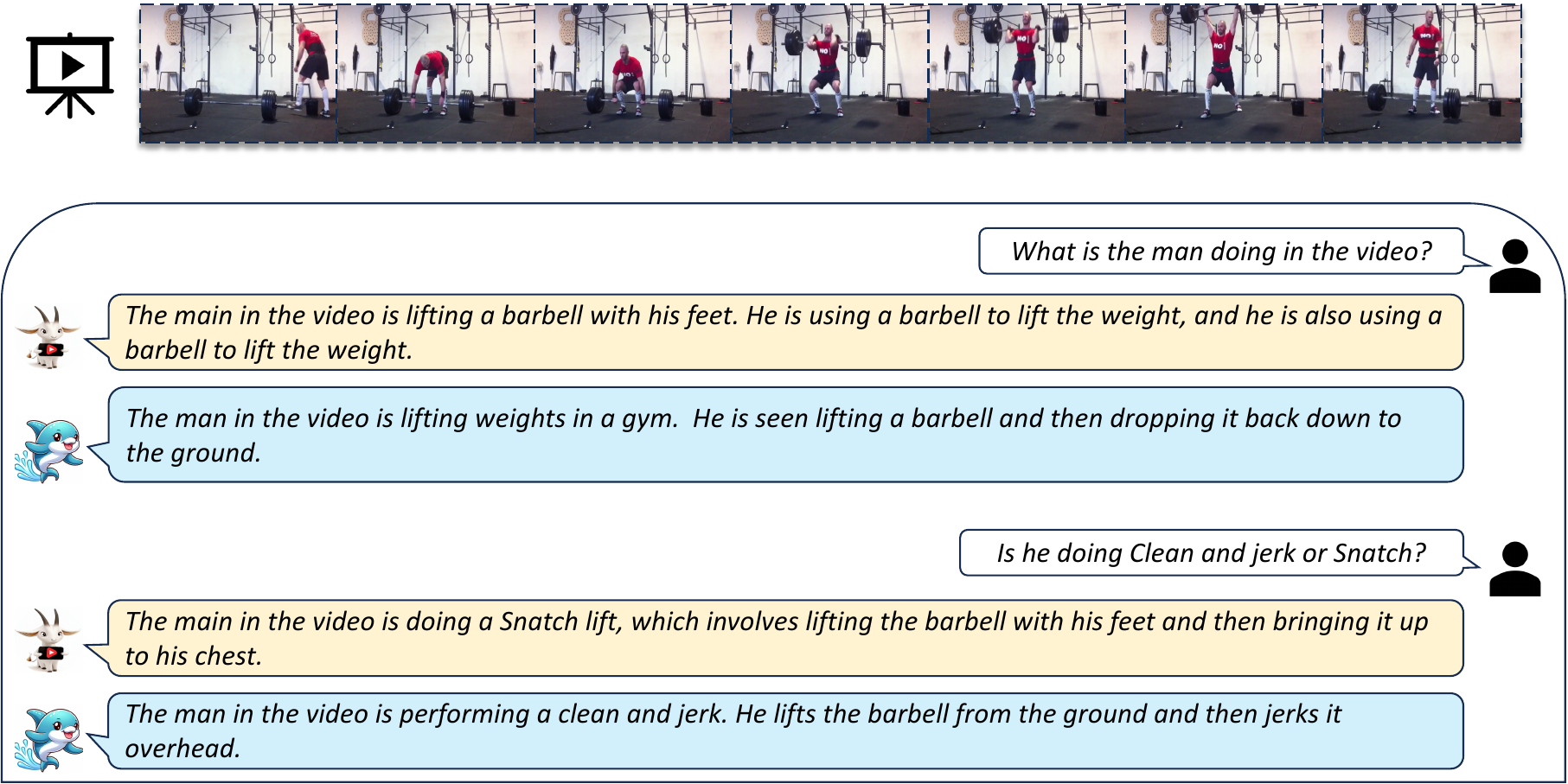}
    \caption{Given a video clip of a man lifting weights, we ask questions on what the man is doing and whether he is doing Clean and Jerk or Snatch. Our \ours answers both the questions correctly. While the baseline Video-ChatGPT~\cite{maaz2023video} generates a repetitive answer to the first question, seeming somewhat chaotic, and fails to discriminate that the man is doing a Clean and Jerk, rather than Snatch. Best viewed in color.}
    \label{fig: demo_weight}
  \end{subfigure}
  \caption{Multi-round conversations of \ours and Video-ChatGPT~\cite{maaz2023video} on two video samples of ActivityNet-200~\cite{caba2015activitynet} dataset.}
  \label{fig: demo}
\end{figure*}

The experiments conducted predominantly address scenarios involving a singular question and answer. However, in practical applications such as Copilot or assistants for industrial products, the capacity for multi-round conversations is crucial for user experience. To evaluate this aspect, we compare the multi-round conversation capabilities of \ours with one of the baselines Video-ChatGPT~\cite{maaz2023video}. As depicted in Fig.~\ref{fig: demo}, \ours demonstrates consistently more accurate and comprehensive conversational abilities compared to Video-ChatGPT. This highlights \ours's potential for industrial applications. More video dialogue examples are provided in the supplementary.

\subsection{Ablation Studies}
We perform ablation studies on removing the components of \ours and altering the options of hyperparameters.

\subsubsection{Ablation of the components of \ours}
We perform ablation studies w.r.t. the components of \ours including the Data Alignment, Feature Alignment, and Prompt Engineering. We conduct experiments on all three Video Question Answering datasets. Tab.~\ref{table: ablation-component} shows that Data Alignment, Feature Alignment, and Prompt Engineering all contribute to zero-shot video QA performance. For all three datasets, the performance of merely adopting Feature Alignment performs better than merely adopting Data Alignment without Prompt Engineering, implying that feature-level learning is comparatively more significant than input data selection for our task.

\begin{table}[tb]
   \centering
   \resizebox{0.9\columnwidth}{!}{
   \begin{tabular}{l c c c c c}
   \toprule
    Datasets & FA& DA &  PE & Accuarcy & Score  \\ \midrule
    \multirow{4}{*}{MSVD-QA} & \XSolidBrush & \Checkmark  & \XSolidBrush & 64.5 & 3.2  \\
    &  \Checkmark & \XSolidBrush  & \XSolidBrush & 70.8&3.5 \\
    & \Checkmark & \Checkmark & \XSolidBrush  & 74.4& 3.7 \\
    & \Checkmark & \Checkmark & \Checkmark & \textbf{74.6} & \textbf{3.7} \\
    \midrule
    \multirow{4}{*}{MSRVTT-QA}& \XSolidBrush  & \Checkmark & \XSolidBrush  & 50.8& 2.9    \\
    &  \Checkmark & \XSolidBrush  & \XSolidBrush & 59.7 & 3.1\\
    & \Checkmark & \Checkmark & \XSolidBrush  & 68.5 & 3.3 \\
    & \Checkmark & \Checkmark & \Checkmark & \textbf{68.6} & \textbf{3.3} \\
    \midrule
    \multirow{4}{*}{Activity Net-QA}&  \XSolidBrush  & \Checkmark & \XSolidBrush  & 44.9& 3.1     \\
    &  \Checkmark & \XSolidBrush  & \XSolidBrush & 47.4&3.1 \\
    & \Checkmark & \Checkmark & \XSolidBrush & 47.7 & 3.3 \\
    &\Checkmark & \Checkmark & \Checkmark & \textbf{48.8} & \textbf{3.3} \\
   \bottomrule
   \end{tabular}
   }
   \vspace{-0.5em}
   \caption{Ablation of the components of \ours. FA, DA, and PE stand for Feature Alignment, Data Alignment, and Prompt Engineering. }
   \vspace{-1em}
   \label{table: ablation-component}
\end{table}

\subsubsection{Ablation of Hyperparameters}\label{sec: ablation-hyper}
We further study the effects of changing the hyperparameter values in our \ours framework. We conduct the ablation studies on the Activity Net-QA testing dataset.

\paragraph{Video Perceiver Depth \& Pretrained Model}
\begin{figure}
  \begin{subfigure}{0.49\linewidth}
    \includegraphics[width=\linewidth]{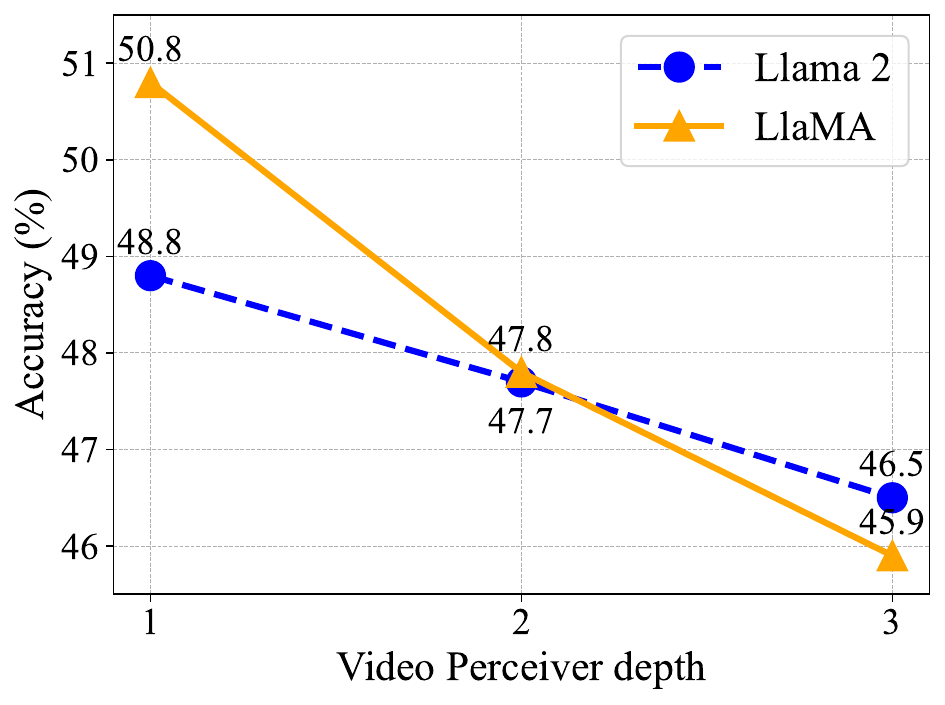}
    \caption{Accuracy w.r.t. depth $p$. }
  \end{subfigure}
  \begin{subfigure}{0.49\linewidth}
    \includegraphics[width=\linewidth]{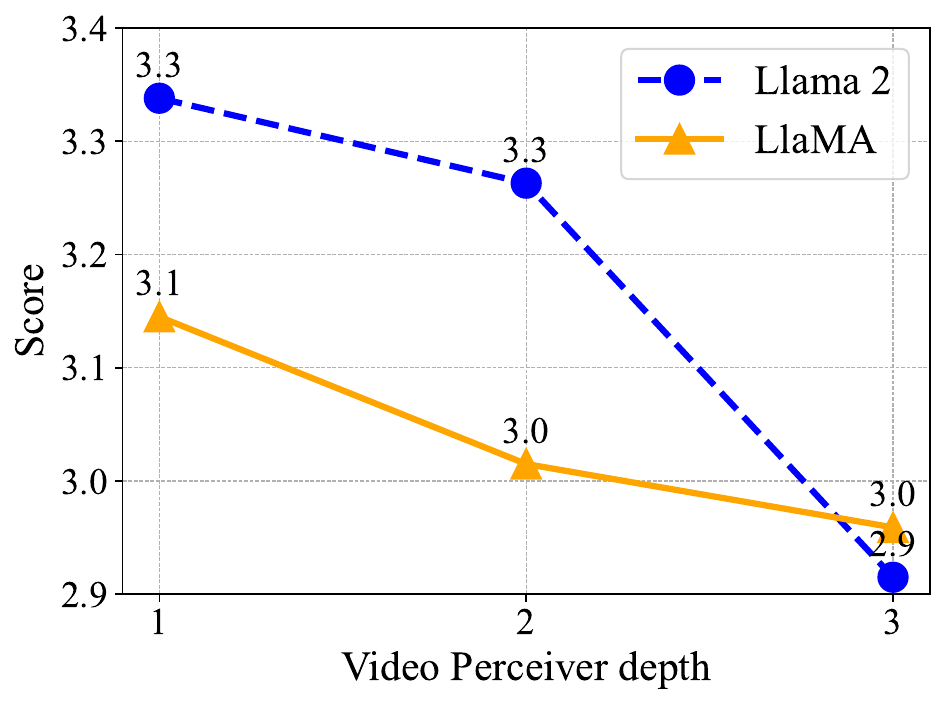}
    \caption{Score w.r.t. depth $p$.}
  \end{subfigure}
  \vspace{-0.5em}
  \caption{Video QA performance on Activity Net-QA~\cite{yu2019activitynet} using pretrained LLama 2~\cite{lamma2_23} and LLaMA~\cite{llama_23}. Best viewed in color.}
  \vspace{-1em}
  \label{fig: ablation-depth-pretrained}
\end{figure}
We try using multiple layers in Video Perceivers and using the LLaMA~\cite{llama_23} model with weight initialization from LLaVA~\cite{liu2023visual}. As illustrated in Fig.~\ref{fig: ablation-depth-pretrained}, the accuracy of \ours drops when the layer number $p$ of the Video Perceiver increases for both LLaMA and Llama 2 backbone. This might largely result from the small training epoch we use and the limited size of training data. For the LLM weights initialized from LLaVA and LLaVA-1.5, we find that the performance gap is not as large as expected, and using LLaMA (LLaVA-1.5) pretrained weights with one layer in Video Perceiver even achieve 50.8 accuracy on Activity Net-QA dataset. On the other hand, models initialized using LLama 2 are obviously more robust to the perceiver depth and are significantly better in relative score evaluation.

\paragraph{Prompt Engineering Design}

\begin{figure}[tb]
  \centering
  \includegraphics[width=0.9\linewidth]{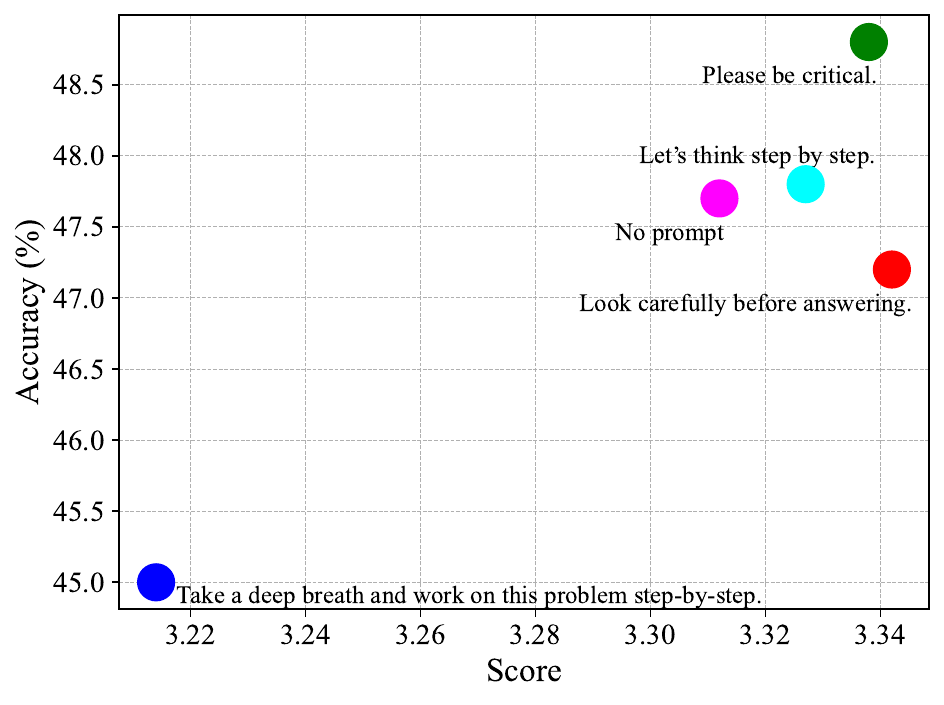}
  \vspace{-1em}
  \caption{Accuracy and score results on Activity Net-QA~\cite{yu2019activitynet} dataset of different prompt designs. Best viewed in color.}
  \label{fig: pe}
  \vspace{-1em}
\end{figure}
We ablate the prompt added before the question. We compare our designed prompt with two popular instruction prompts in the NLP filed: ``Take a deep breath and work on this problem step-by-step.''~\cite{yang2023large} and ``Let's think step by step.''~\cite{kojima2022large}. We also compare with another prompt ``Look carefully before answering.'' and indicate the performance when not adding a prompt. From the accuracy and score results shown in Fig.~\ref{fig: pe}, we can draw the conclusion that our designed prompt, ``Please be critical'', performs the best with both the highest accuracy and the highest score. ``Let's think step by step.''~\cite{kojima2022large} improves the performance slightly while ``Take a deep breath and work on this problem step-by-step.''~\cite{yang2023large} degrades the performance.

\section{Conclusion}

Our proposed \ours represents a significant stride in the field of video understanding. By moving away from traditional frame sampling methods and adopting a CLIP-score guided technique, we have achieved a more nuanced and effective integration of video frame and text data. The innovative combination of a trainable video perceiver with a visual-query transformer mechanism allows for a dynamic interplay between video features and input questions, further augmented by the strategic use of prompts. The results from our studies clearly demonstrate that \ours not only excels in zero-shot video question-answering tasks but also in generating coherent and contextually rich multi-turn video dialogues. Our proposed \ours therefore sets a new standard for LLM-based video understanding.

\clearpage

\clearpage
\setcounter{page}{1}

\section{Raw Videos of Sec.~\ref{sec: multi-round}}
We supplement the raw videos of the two examples in Sec.~\ref{sec: multi-round}, namely \href{https://www.dropbox.com/scl/fi/jx5ez8628o3ji4t90k8nj/multi_round_example_1.mp4?rlkey=fdo3d8gwrr1ytnyvkjt4g8ikc&dl=0}{``multi$\_$round$\_$example$\_$1.mp4"} and \href{https://www.dropbox.com/scl/fi/v28ee1mzwvezasspka0zg/multi_round_example_2.mp4?rlkey=komv36os0qfia1zy56kleoetk&dl=0}{``multi$\_$round$\_$example$\_$2.mp4"}. They are chosen from the test set of ActivityNet-200~\cite{caba2015activitynet} dataset.

\section{\ours Assistant Demo}

We also provide a video demo recording of our \ours Assistant on Gradio~\cite{abid2019gradio} (\href{https://www.dropbox.com/scl/fi/zrz3qei6wsrfdnm68by81/VaQuitA_demo.mp4?rlkey=150nakspz8k2mlcnjwlc1s1e2&dl=0}{``VaQuitA$\_$demo.mp4"}). The three example videos are chosen from the test set of TGIF~\cite{li2016tgif}, Social-IQ 2.0 and ActivityNet-200~\cite{caba2015activitynet} datasets. The videos are about a boy falling down the skateboard on a ramp, a doctor and patient talking to each ohter in the hospital and a man shaving himself in the bathroom, respectively. We show in our demo recording that the \ours Assistant is able to generate high-quality multi-round conversations at a high responding speed. It is able to precisely summarize the content of video, identify the relationships between characters and events, and pinpoint locations.

\section{Test-time Data Alignment}
We conduct additional experiment using our proposed sampling approach in Sec.~\ref{sec: da} during the inference stage. We use the Video-ChatGPT~\cite{maaz2023video} trained model and only change the sampling way in inference. The baseline is uniform sampling. Given a video clip of MSNBC news report (\href{https://www.dropbox.com/scl/fi/ctml8hlsh4d04bhbrof2g/test_time_da_example.mp4?rlkey=abuz5ymc7mhcgri4yklvkx9l5&dl=0}{``test$\_$time$\_$da$\_$example.mp4"}),  we ask a video question: ``During the movie, there is a video clip with flying animals. What is the flying animal, bird or bat?"for $3$ independent times. The correct answer is ``bat", which corresponds to 2:16-2:22 time stamp of the video. For uniform sampling, the model answers: ``The flying animal in the video is a bird." for $3$ times, which is wrong; for our proposed sampling method, the model answers: ``The flying animal in the video is a bat" for $3$ times, which is correct. 

This superiority of our Data Alignment module mainly results from the CLIP Feature Similarity-based Frame Selection component, which is verified by checking the selected frames. We supplement the directories of the sampled frame of uniform sampling and our data alignment sampling method. The sampled frames using uniform sampling are stored under directory \href{https://www.dropbox.com/scl/fo/4gz92rfhm5y2hv9i86wrv/h?rlkey=iq94yh3f0014t464keuyun23d&dl=0}{``uniform$\_$sampled$\_$frames"} and the sampled frames using our proposed sampling method are under directory \href{https://www.dropbox.com/scl/fo/z1rnps07j6monf77bnwye/h?rlkey=jkm454gd94gsoltnbcy5ls8by&dl=0}{``ours$\_$sampled$\_$frames"}. We can see that the uniform sampling only samples one frame (``frame$\_$4223.jpg") related to the question, while our proposed sampling method samples 13 related frames (``frame4197.jpg", ``frame4198.jpg", ``frame4201.jpg", ``frame4206.jpg", ``frame4207.jpg", ``frame4247.jpg", ``frame4280.jpg", ``frame4281.jpg", ``frame4282.jpg", ``frame4287.jpg", ``frame4197.jpg", ``frame4288.jpg", ``frame4289.jpg", ``frame4304.jpg"). Since our sampling method samples more frames corresponding to the question, the model can answer more correctly, which reflects the effect of Data Alignment in the inference phase.

{
    \small
    \bibliographystyle{ieeenat_fullname}
    \bibliography{vaquita_arxiv}
}

\end{document}